\documentclass[11pt,a4paper]{article}

\providecommand{\corresp}{\noindent}

\usepackage[utf8]{inputenc}
\usepackage[top=2cm, bottom=2cm, left=1.5cm, right=1.5cm]{geometry}

\usepackage{ragged2e}

\usepackage{amsmath}
\usepackage{amssymb}
\usepackage{physics}

\usepackage{graphicx}
\usepackage{subcaption}
\usepackage{float}

\usepackage{booktabs}
\usepackage{multirow}
\usepackage{array}
\usepackage{tabularx}
\newcolumntype{L}{>{\raggedright\arraybackslash}X} 

\newcolumntype{F}[1]{>{\raggedright\arraybackslash}p{#1}}

\usepackage{caption}
\usepackage{needspace}

\makeatletter

\renewenvironment{table*}[1][htbp]{%
  \@dblfloat{table}[#1]%
  \needspace{5\baselineskip}%
}{%
  \end@dblfloat
}

\renewenvironment{figure*}[1][htbp]{%
  \@dblfloat{figure}[#1]%
  \needspace{5\baselineskip}%
}{%
  \end@dblfloat
}
\makeatother

\renewcommand{\arraystretch}{0.75} 
\newcommand{\tablesize}{\normalsize} 
\newcolumntype{Y}{>{\centering\arraybackslash}X} 

\usepackage{multicol}

\usepackage{listings}
\usepackage{xcolor}

\usepackage{cite}
\usepackage{hyperref}

\usepackage{titlesec}
\titleformat{\section}
  {\normalfont\large\bfseries}{\thesection}{0.5em}{}
\titleformat{\subsection}
  {\normalfont\normalsize\bfseries}{\thesubsection}{0.5em}{}

\titlespacing*{\section}{0pt}{8pt plus 2pt minus 2pt}{4pt plus 2pt minus 2pt}
\titlespacing*{\subsection}{0pt}{6pt plus 2pt minus 2pt}{3pt plus 2pt minus 2pt}

\usepackage{fancyhdr}
\title{HBGSA: Hydrogen Bond Graph with Self-Attention for Drug-Target Binding Affinity Prediction}

\author{
Junxiao Kong,$^{1,\dagger}$ 
Chupei Tang,$^{1,\dagger}$ 
Di Wang,$^{1}$ 
Jixiu Zhai,$^{1}$ 
Yi He,$^{1}$ 
Moyu Tang,$^{1}$ 
and Tianchi Lu$^{2,*}$
}

\date{}

\begin{document}


\maketitle

\noindent
$^{1}$School of Mathematics and Statistics, Lanzhou University, 222 South Tianshui Road, 730000, Gansu, China

\noindent
$^{2}$Department of Computer Science, City University of Hong Kong, 83 Tat Chee Avenue, Kowloon Tong, Hong Kong 999077, China

\noindent
$^{\dagger}$These authors contributed equally to this work.

\noindent
\corresp$\ast${Corresponding author. \href{mailto:tianchilu4-c@my.cityu.edu.hk}{tianchilu4-c@my.cityu.edu.hk} (T.L.) Department of Computer Science, City University of Hong Kong, 83 Tat Chee Avenue, Kowloon Tong, Hong Kong 999077, China, Tel: +86-13239620274}

\vspace{0.5em}


\noindent
\textbf{Abstract}

\noindent
Accurate prediction of drug-target binding affinity accelerates drug discovery by prioritizing compounds for experimental validation. Current methods face three limitations: sequence-based approaches discard spatial geometric constraints, structure-based methods fail to exploit hydrogen bond features, and conventional loss functions neglect prediction-target correlation—a key factor for identifying high-affinity compounds in virtual screening. We developed HBGSA (Hydrogen Bond Graph with Self-Attention), a 3.06M-parameter model that encodes hydrogen bond spatial features. HBGSA uses graph neural networks to model hydrogen bond spatial topology with self-attention enhancement and Pearson correlation loss. Experimental results on PDBbind Core Set and CSAR-HiQ dataset demonstrate that HBGSA outperforms baseline methods with strong generalization capability. Ablation studies confirm the effectiveness of hydrogen bond modeling and Pearson correlation loss.

\vspace{0.5em}


\vspace{1em}

\begin{multicols}{2}

\section{Introduction}
\justifying

Drug discovery faces high experimental costs\cite{dimasi2016}. Computational models prioritizing high-affinity candidates before synthesis reduce experimental burden. Drug-target affinity (DTA) prediction is central to this virtual screening workflow.

Existing DTA methods fall into two categories. Sequence-based methods like DeepDTA\cite{deepdta} and DeepDTAF\cite{deepdtaf} use 1D convolutions on sequences and SMILES, but discard spatial structural information.

Structure-based methods like GraphDTA\cite{graphdta}, MMPD-DTA\cite{mmpd-dta}, and ML-PLA\cite{mlpla} construct molecular graphs from spatial coordinates. However, most methods define edges using fixed distance thresholds (e.g., 6.0 Å in MMPD-DTA), treating all atomic contacts equally. This approach fails to distinguish hydrogen bonds from van der Waals contacts. Recent works like PLIG\cite{plig} and IGN\cite{ign} encode interaction types as node features, but lack explicit modeling of hydrogen bond spatial patterns. Existing graph neural networks treat atoms as isotropic nodes, unable to encode hydrogen bond spatial distributions or model cooperative effects of hydrogen bond networks. Furthermore, most methods use MSE loss, which optimizes absolute error but neglects prediction-target correlation—a key factor for identifying high-affinity compounds in virtual screening.

HBGSA addresses these limitations through three innovations:

(i) \textbf{Hydrogen Bond Graph Architecture}: Graph neural networks model spatial topology and cooperative effects of hydrogen bond interactions.

(ii) \textbf{Pearson Correlation Loss}: Combining SmoothL1 regression loss with Pearson correlation loss optimizes both absolute prediction accuracy and prediction-target correlation, enhancing the model's ability to identify high-affinity compounds.

(iii) \textbf{Self-Attention Enhancement}: Self-attention modules in sequence and SMILES encoders improve feature utilization and representation quality.

\textbf{Paper Organization:} Section 2 reviews sequence-based and structure-based methods, optimization objectives, and attention mechanisms. Section 3 details datasets and metrics (3.1-3.2), then presents the HBGSA architecture including multimodal encoders (3.3-3.4), graph-based hydrogen bond modeling (3.5), and hybrid loss optimization (3.6). Section 4 presents results on PDBbind v2016 Core and CSAR-HiQ, with ablation studies validating each component. Section 5 discusses hydrogen bond density analysis, correlation-based optimization, and model limitations.

\section{Related Work}
\justifying

\subsection{Sequence-based Approaches}
\justifying

Sequence-based methods encode proteins as amino acid strings and ligands as SMILES. DeepDTA (2018)\cite{deepdta} pioneered this approach using 1D CNNs on one-hot encoded sequences. WideDTA\cite{widedta} added ligand maximum common substructures and protein motifs as auxiliary features. DeepDTAF (2021)\cite{deepdtaf} focused on binding pockets and introduced dilated convolutions to model long-range dependencies, substantially improving accuracy. MMPD-DTA\cite{mmpd-dta} explored Transformer architectures\cite{attention} for richer contextual representations.

These methods treat proteins as 1D strings, discarding spatial topology. This loses geometric relationships between residues, limiting physical interpretability of binding mechanisms.

\subsection{Structure-based Graph Learning}
\justifying

Structure-based methods encode spatial molecular geometry. Traditional approaches like RF-Score and ECIF use hand-crafted features. Structure-based CNN methods emerged with Pafnucy\cite{pafnucy}, OnionNet, and DeepAtom, which voxelize protein-ligand complexes. GraphDTA\cite{graphdta} represents ligands as molecular graphs. IGN\cite{ign}, SIGN, and SGADN construct interaction graphs containing both protein and ligand atoms. MMPD-DTA\cite{mmpd-dta} proposes pocket-drug graphs combined with GraphSAGE for neighbor aggregation. ML-PLA\cite{mlpla} uses heterogeneous graph neural networks to model protein microenvironments.

Geometric learning methods include PLIG, EGNN, TankBind\cite{tankbind}, and GIGN. Advanced multimodal approaches like HAC-Net, MMDTA, MBP, and UAMRL combine multiple representations and pre-training strategies. CAPLA and KSM represent recent advances in this field.

Graph construction methods like MMPD-DTA and ML-PLA typically use fixed distance thresholds. This treats all atomic contacts equally, failing to distinguish hydrogen bonds from van der Waals contacts. HBGSA addresses this by explicitly modeling hydrogen bond spatial features (Section 3.5).

\subsection{Optimization Objectives}
\justifying

Most methods (DeepDTAF\cite{deepdtaf}, GraphDTA\cite{graphdta}, MMPD-DTA\cite{mmpd-dta}) use mean squared error (MSE) loss, which optimizes absolute prediction error. However, virtual screening requires strong correlation between predictions and true affinities to reliably identify high-affinity compounds. Some recent works explore alternative objectives: TankBind\cite{tankbind} introduces contrastive loss, while MBP uses pairwise ranking loss to handle label noise.

Direct optimization of global correlation (Pearson coefficient\cite{pearson}) in DTA regression remains underexplored. Section 3.6 details our Pearson correlation loss.

\subsection{Attention Mechanisms}
\justifying

Attention mechanisms dynamically weight key regions. AttentionDTA and AttentiveFP enable models to focus on atoms or residues contributing most to binding. These mechanisms apply primarily to sequence feature extraction, using self-attention to model long-range dependencies and enhance feature expressiveness.

\begin{figure*}[!t]
\centering
\includegraphics[width=0.95\textwidth]{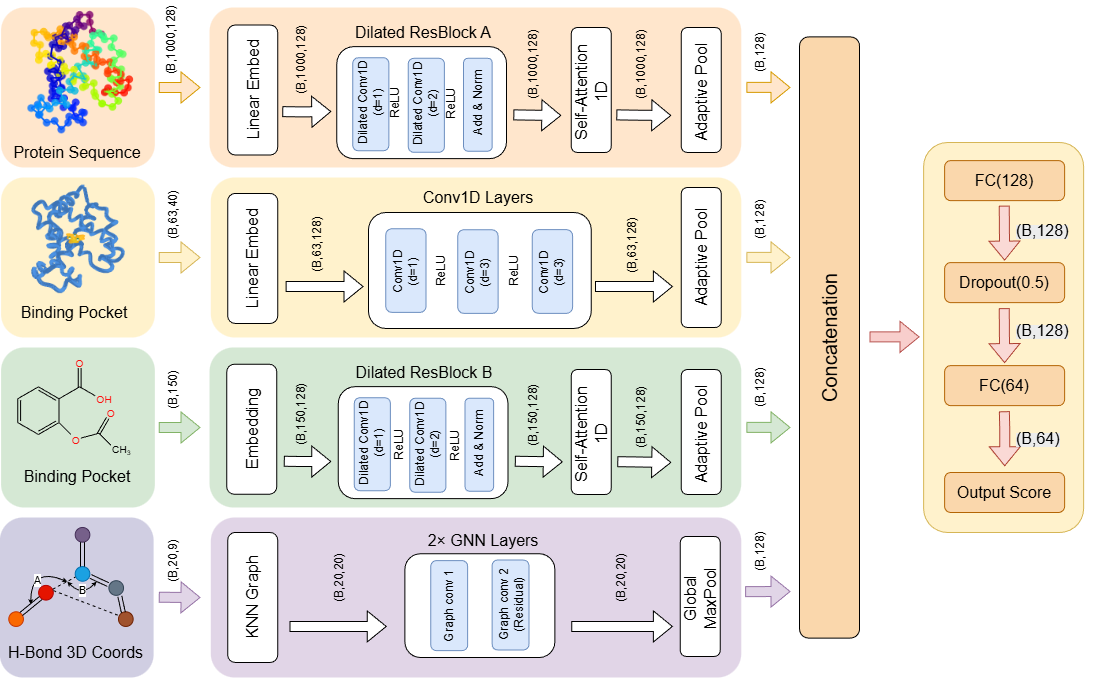}
\caption{Overall architecture of the HBGSA model.}
\label{fig:overall_architecture}
\end{figure*}

\section{Methods}
\justifying

\subsection{Datasets}
\justifying

We use PDBbind v2016\cite{pdbbind} as the primary benchmark, which includes experimentally verified protein-ligand binding affinity from Protein Data Bank. We focus on three subsets: General Set (originally 9,226 complexes), Refined Set (originally 4,057 complexes), and Core Set 2016 (290 complexes as the standard test benchmark).

To prevent data overlap, we perform data cleaning: (1) remove 290 Core Set complexes from Refined Set; (2) remove 82 validation and 5 training complexes to maintain consistency with DeepDTAF\cite{deepdtaf} and Pafnucy\cite{pafnucy}. After cleaning, General Set contains 9,221 complexes and Refined Set contains 3,685 complexes.

Data partition: Test Set uses Core Set 2016 (290 complexes); Validation Set randomly samples 1,000 from cleaned Refined Set; Training Set combines General Set (9,221) with remaining Refined Set (2,685), totaling 11,906 complexes.

We also evaluate on CSAR-HiQ\cite{csar}, merging two subsets (176 and 167 complexes) and removing overlaps with PDBbind Refined Set, yielding 135 independent test complexes following MBP\cite{mbp} and SGADN\cite{sgadn} protocols.

Maximum sequence lengths cover approximately 90\% of samples: 1,000 residues for protein sequences, 63 for binding pockets, and 150 characters for SMILES (consistent with DeepDTAF\cite{deepdtaf}). For hydrogen bonds, we select top $N=20$ strongest bonds per complex, covering 96.61\% of training samples (Section 4.4). 

\subsection{Evaluation Metrics}
\justifying

We evaluate model performance using four standard metrics (detailed calculation methods in Supplementary File Section 1). Let $y_i$ denote the true binding affinity and $\hat{y}_i$ the predicted value for the $i$-th sample, with $N$ total samples.

\textbf{Root Mean Squared Error (RMSE)} measures prediction accuracy:
\begin{equation}
\text{RMSE} = \sqrt{\frac{1}{N}\sum_{i=1}^{N}(y_i - \hat{y}_i)^2}
\end{equation}

\textbf{Mean Absolute Error (MAE)} measures average absolute deviation:
\begin{equation}
\text{MAE} = \frac{1}{N}\sum_{i=1}^{N}|y_i - \hat{y}_i|
\end{equation}

\textbf{Pearson Correlation Coefficient (Pearson R)} measures linear correlation:
\begin{equation}
r = \frac{\sum_{i=1}^{N}(y_i - \bar{y})(\hat{y}_i - \bar{\hat{y}})}{\sqrt{\sum_{i=1}^{N}(y_i - \bar{y})^2}\sqrt{\sum_{i=1}^{N}(\hat{y}_i - \bar{\hat{y}})^2}}
\end{equation}
where $\bar{y}$ and $\bar{\hat{y}}$ are the means of true and predicted values.

\textbf{Concordance Index (CI)} measures prediction monotonicity:
\begin{equation}
\text{CI} = \frac{1}{Z}\sum_{i,j:y_i>y_j}\mathbb{1}(\hat{y}_i > \hat{y}_j)
\end{equation}
where $Z$ is the number of comparable pairs and $\mathbb{1}(\cdot)$ is the indicator function.

\subsection{Architecture Overview}
\justifying

HBGSA is a compact 3.06M-parameter model that integrates four input modalities through parallel encoders (Figure \ref{fig:overall_architecture}; detailed parameter breakdown in Supplementary Table 2): protein sequences, binding pockets, ligand SMILES, and hydrogen bond spatial coordinates. The protein and SMILES encoders use dilated convolutions with self-attention; the pocket encoder uses standard 1D convolutions; the hydrogen bond encoder uses graph neural networks to model spatial topology. After adaptive pooling, the four feature vectors are concatenated and passed through a fully connected predictor with dropout to output binding affinity.

\subsection{Multi-modal Feature Extraction}
\justifying

Following DeepDTAF\cite{deepdtaf}, HBGSA processes three sequence modalities: global protein sequences, local binding pockets, and ligand SMILES. The architecture of protein sequence and binding pocket encoders is adapted from DeepDTAF\cite{deepdtaf}. Protein sequences use 40-dimensional physicochemical property vectors encoding hydrophobicity, charge, polarity, and secondary structure propensity. All inputs are projected to a unified 128-dimensional feature space.

For protein sequences and ligand SMILES, we use multi-scale dilated convolutions to model long-range dependencies. Dilated residual blocks stack 1D convolutional layers, with residual connections\cite{resnet} and layer normalization to prevent gradient vanishing. For the shorter binding pocket branch, standard 1D convolutional layers (kernel size 3) encode local features.

To identify key pharmacophores and residues, we add 1D self-attention modules at the end of sequence and SMILES branches, and use adaptive max pooling to generate fixed-dimensional vectors $\mathbf{v}_{seq}, \mathbf{v}_{pkt}, \mathbf{v}_{smi} \in \mathbb{R}^{128}$ for multimodal fusion (Figure \ref{fig:attention_mechanism}).

\begin{center}
\includegraphics[width=1.00\columnwidth]{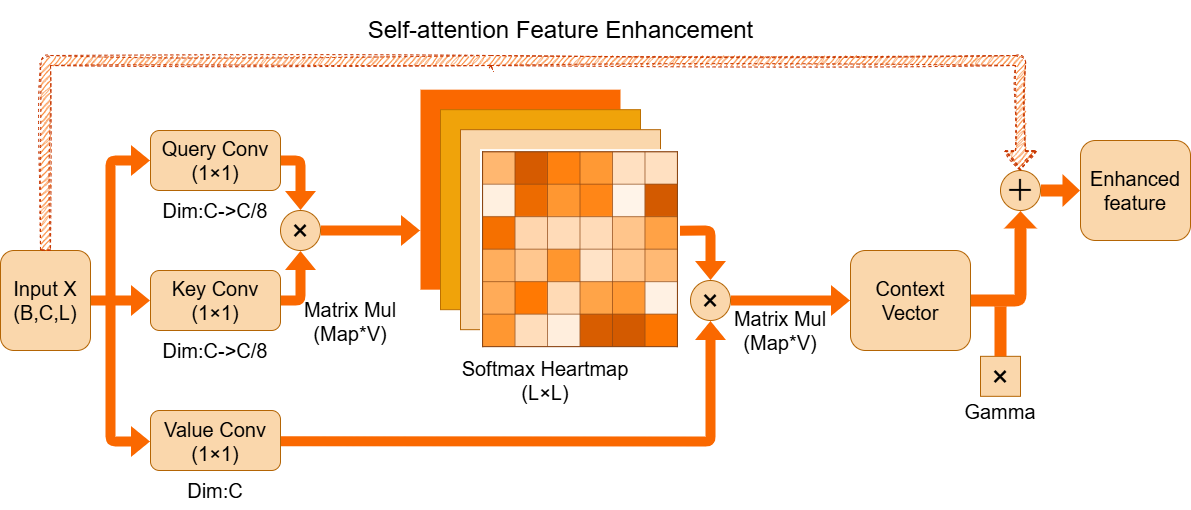}
\captionof{figure}{Self-attention feature enhancement module.}
\label{fig:attention_mechanism}
\end{center}

\subsection{Graph-Based Interaction Learning}
\justifying

The hydrogen bond module is the core innovation of HBGSA. Most methods treat interactions as 1D sequences or simple distance contacts, failing to model spatial topology and cooperative effects of hydrogen bonds. HBGSA uses graph neural networks to capture spatial relationships among hydrogen bonds.

\textbf{Hydrogen Bond Extraction and Selection.} HBGSA is designed as a multimodal model that can leverage 3D structural information when available. When protein-ligand complex structures are available (from crystallography or docking), we extract hydrogen bonds using PyMOL's distance command (mode=2), which identifies them based on standard geometric criteria: donor-acceptor distance $d \leq 3.5$ Å, D-H...A angle $\geq 120°$, and polar atom types (N, O, S). This definition follows the IUPAC hydrogen bond standard\cite{hbond}. For each complex, we rank all identified hydrogen bonds by distance and select the top $N=20$ strongest, which covers 96.61\% of samples in the training set. Complexes with fewer than 20 bonds are zero-padded. When 3D structures are unavailable, the model can still make predictions using only sequence and SMILES information by setting hydrogen bond inputs to zero vectors, though with reduced accuracy.

Each hydrogen bond is represented by a 9-dimensional coordinate vector: protein end $\mathbf{p}_i \in \mathbb{R}^3$, ligand end $\mathbf{l}_i \in \mathbb{R}^3$, and midpoint $\mathbf{m}_i = (\mathbf{p}_i + \mathbf{l}_i)/2 \in \mathbb{R}^3$. The global hydrogen bond information $\mathbf{x}_{hb} \in \mathbb{R}^{180}$ is reshaped into $\mathbf{X}_{hb} \in \mathbb{R}^{20 \times 9}$ as input to the graph encoder.

\begin{figure*}[!t]
\centering
\includegraphics[width=0.85\textwidth]{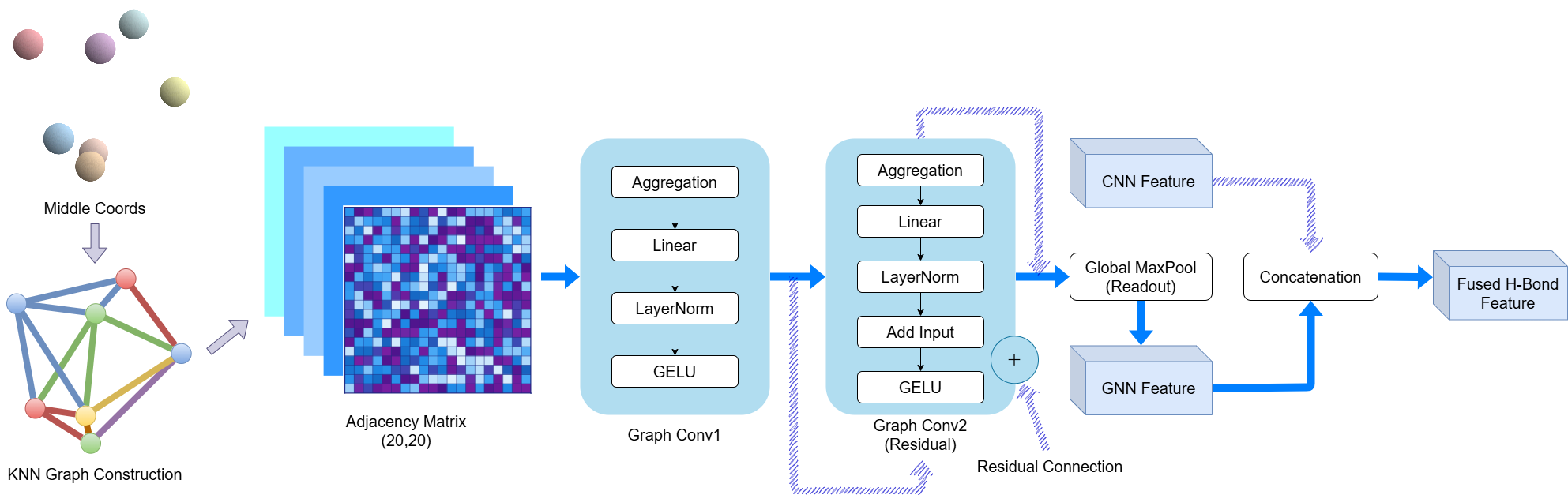}
\caption{Hydrogen bond graph neural network encoder.}
\label{fig:gnn_pathway}
\end{figure*}

\textbf{GNN Encoder} models spatial geometric topological relationships among hydrogen bonds through graph neural networks (Figure \ref{fig:gnn_pathway}). Treating the 20 hydrogen bonds as graph nodes, node features are first embedded: $\mathbf{h}_i^{(0)} = \text{Linear}(\mathbf{X}_{hb,i}) \in \mathbb{R}^{128}$. Spatial relationships are encoded through pairwise distances calculated using midpoint coordinates $d_{ij} = \|\mathbf{m}_i - \mathbf{m}_j\|_2$. A dynamic KNN graph is constructed by selecting the $k=5$ nearest spatial neighbors for each bond, forming adjacency matrix $\mathbf{A}$.

Graph convolutional layers aggregate neighbor information through the adjacency matrix. The first graph convolution:
\begin{equation}
\mathbf{H}^{(1)} = \text{GELU}(\text{LayerNorm}(\text{Linear}(\mathbf{A} \cdot \mathbf{H}^{(0)})))
\end{equation}
where $\mathbf{A} \in \mathbb{R}^{20 \times 20}$ is the KNN adjacency matrix and $\mathbf{H}^{(0)} \in \mathbb{R}^{20 \times 128}$ is the node feature matrix. The second layer adds residual connections:
\begin{equation}
\mathbf{H}^{(2)} = \text{GELU}(\mathbf{H}^{(1)} + \text{LayerNorm}(\text{Linear}(\mathbf{A} \cdot \mathbf{H}^{(1)})))
\end{equation}
Finally, global max pooling yields $\mathbf{s}_{hb} = \max(\mathbf{H}^{(2)}, \text{dim}=0) \in \mathbb{R}^{128}$.

\subsection{Prediction and Optimization}
\justifying

We concatenate features from all branches into a unified representation:
\begin{equation}
\mathbf{s}_{cat} = [\mathbf{s}_{seq}; \mathbf{s}_{pkt}; \mathbf{s}_{smi}; \mathbf{s}_{hb}] \in \mathbb{R}^{512}
\end{equation}
A three-layer fully connected network progressively downsamples the fused features:
\begin{align}
\mathbf{h}_1 &= \text{PReLU}(\text{Dropout}(\text{Linear}_1(\mathbf{s}_{cat}))) \\
\mathbf{h}_2 &= \text{PReLU}(\text{Dropout}(\text{Linear}_2(\mathbf{h}_1))) \\
\mathbf{y} &= \text{Linear}_3(\mathbf{h}_2)
\end{align}
where layer dimensions are $512 \to 128 \to 64 \to 1$, with dropout ratio 0.5.

We design a hybrid loss function to optimize both absolute prediction accuracy and prediction-target correlation:
\begin{equation}
\mathcal{L} = \mathcal{L}_{reg} + \lambda \mathcal{L}_{pearson}
\end{equation}

where $\mathcal{L}_{reg}$ is the SmoothL1 loss (Huber loss), robust to outliers:
\begin{equation}
\mathcal{L}_{reg} = \frac{1}{N}\sum_{i=1}^{N} \begin{cases} 
\frac{1}{2}(y_i - \hat{y}_i)^2 & \text{if } |y_i - \hat{y}_i| \leq 1 \\
|y_i - \hat{y}_i| - \frac{1}{2} & \text{otherwise}
\end{cases}
\end{equation}
This loss uses squared loss for small errors (sensitive to small errors) and absolute value loss for large errors (robust to outliers).

$\mathcal{L}_{pearson}$ is our Pearson correlation coefficient loss, defined as $\mathcal{L}_{pearson} = 1 - r_{pearson}$, where $r_{pearson}$ is the Pearson correlation between predicted and true values. This loss calculates correlation after centering predicted and true values. When $r_{pearson} = 1$ (perfect positive correlation), $\mathcal{L}_{pearson} = 0$; when $r_{pearson} = 0$ (no correlation), $\mathcal{L}_{pearson} = 1$.

This loss function provides three benefits: (i) \textbf{Correlation Optimization}: forces the model to learn correct relative relationships rather than just minimizing absolute errors. Strong correlation ensures that compounds with higher predictions indeed have higher true values, which is crucial for prioritizing candidates in virtual screening;(ii) \textbf{Robustness to Systematic Bias}: Pearson correlation is invariant to linear transformations of predictions (e.g., constant shifts, Pearson loss remains small);(iii) \textbf{Improved Monotonicity}: Optimizing correlation naturally improves monotonicity—the tendency for higher predictions to match higher true values. Our ablation study demonstrates that the optimal $\lambda$ balances correlation and absolute accuracy\mbox{ (see Section 4.2)}.

\needspace{3\baselineskip}
\section{Results}
\justifying

\subsection{Performance Comparison and Generalization}
\justifying

Table \ref{tab:core_comparison} compares HBGSA with baseline methods on PDBbind Core Set.

\begin{table*}[!t]
\centering
\renewcommand{\arraystretch}{0.8}
\tablesize
\caption{Performance on PDBbind v2016 Core Set}
\label{tab:core_comparison}
\begin{tabularx}{\linewidth}{@{} l Y Y Y Y @{}}
\toprule
\textbf{Method} & \textbf{RMSE} $\downarrow$ & \textbf{MAE} $\downarrow$ & \textbf{Pearson R} $\uparrow$ & \textbf{CI} $\uparrow$ \\
\midrule
\multicolumn{5}{@{}l}{\textit{Traditional ML \& Sequence/2D-based Methods}} \\
RF-Score (2010) & 1.468 & 1.173 & 0.767 & - \\
ECIF (2021) & 1.362 & 1.080 & 0.803 & - \\
DeepDTA (2018) & 1.443 & 1.148 & 0.749 & 0.771 \\
DeepDTAF (2021) & 1.355 & 1.073 & 0.789 & 0.799 \\
GraphDTA (2021) & 1.658 & 1.308 & 0.653 & - \\
CAPLA (2023) & 1.200 & 0.966 & 0.843 & 0.820 \\
\midrule
\multicolumn{5}{@{}l}{\textit{Structure-based Methods (CNN/Contact)}} \\
Pafnucy (2018) & 1.418 & 1.129 & 0.775 & 0.789 \\
OnionNet (2019) & 1.278 & 0.984 & 0.816 & - \\
DeepAtom (2019) & 1.232 & 0.904 & 0.831 & - \\
\midrule
\multicolumn{5}{@{}l}{\textit{Geometric \& Graph-based Methods (GNNs)}} \\
PLIG (2022) & 1.646 & 1.316 & 0.664 & - \\
EGNN (2021) & 1.377 & 1.076 & 0.778 & - \\
TankBind (2022) & 1.346 & 1.070 & 0.726 & - \\
SGADN (2023) & 1.342 & 1.045 & 0.800 & - \\
SIGN (2021) & 1.316 & 1.027 & 0.797 & 0.754 \\
KSM (2025) & 1.288 & 0.975 & 0.813 & - \\
IGN (2021) & 1.220 & 0.940 & 0.837 & - \\
GIGN (2023) & 1.190 & 0.933 & 0.840 & 0.816 \\
\midrule
\multicolumn{5}{@{}l}{\textit{Advanced Multimodal \& Pre-training Methods}} \\
HAC-Net (2023) & 1.205 & 0.971 & 0.846 & - \\
MMDTA (2024) & 1.166 & 0.921 & 0.844 & 0.822 \\
ML-PLA (2023) & 1.179 & 0.892 & 0.845 & 0.827 \\
MBP (2024) & 1.263 & 0.999 & 0.825 & - \\
UAMRL (2025) & 1.121 & 0.869 & 0.863 & 0.834 \\
\midrule
\textbf{HBGSA (Ours)} & \textbf{1.099} & \textbf{0.752} & \textbf{0.865} & \textbf{0.850} \\
\bottomrule
\end{tabularx}
\end{table*}

HBGSA achieves state-of-the-art performance across all metrics, with RMSE=1.099 and Pearson R=0.865 on PDBbind Core Set.

Table \ref{tab:csar_comparison} shows performance on CSAR-HiQ, an independent test set with 135 complexes.

\begin{table*}[!t]
\centering
\renewcommand{\arraystretch}{0.8}
\tablesize
\caption{Performance on CSAR-HiQ dataset}
\label{tab:csar_comparison}
\begin{tabularx}{\linewidth}{@{} l Y Y Y Y @{}}
\toprule
\textbf{Method} & \textbf{RMSE} $\downarrow$ & \textbf{MAE} $\downarrow$ & \textbf{Pearson R} $\uparrow$ & \textbf{CI} $\uparrow$ \\
\midrule
\multicolumn{5}{@{}l}{\textit{Traditional ML \& Sequence/2D-based Methods}} \\
RF-Score (2010) & 1.947 & 1.466 & 0.723 & - \\
DeepDTAF (2021) & 1.974 & 1.601 & 0.146 & 0.670 \\
GraphDTA (2021) & 2.023 & 1.530 & 0.577 & - \\
\midrule
\multicolumn{5}{@{}l}{\textit{Structure-based Methods (CNN/Contact)}} \\
Pafnucy (2018) & 1.939 & 1.562 & 0.686 & - \\
OnionNet (2019) & 1.927 & 1.471 & 0.690 & - \\
\midrule
\multicolumn{5}{@{}l}{\textit{Geometric \& Graph-based Methods (GNNs)}} \\
IGN (2021) & 1.864 & 1.455 & 0.741 & 0.676 \\
SIGN (2021) & 1.618 & 1.238 & 0.772 & - \\
SGADN (2023) & 1.619 & 1.261 & 0.772 & - \\
\midrule
\multicolumn{5}{@{}l}{\textit{Advanced Multimodal \& Pre-training Methods}} \\
MBP (2024) & 1.624 & 1.240 & 0.791 & - \\
\midrule
\textbf{HBGSA (Ours)} & \textbf{1.603} & \textbf{1.283} & \textbf{0.808} & \textbf{0.809} \\
\bottomrule
\end{tabularx}
\end{table*}

HBGSA achieves RMSE=1.603 and Pearson R=0.808, demonstrating strong generalization.

\subsection{Ablation Studies}
\justifying

Table \ref{tab:ablation_loss} shows that hybrid loss (SmoothL1 + Pearson, denoted as S1+P) achieves both strong absolute prediction accuracy (RMSE=1.099) and high correlation (Pearson R=0.865). As a result, the model also exhibits improved monotonicity (CI=0.850), outperforming SmoothL1 alone (S1).

\begin{center}
\tablesize
\captionof{table}{Ablation study on loss functions}
\label{tab:ablation_loss}
\begin{tabularx}{\columnwidth}{@{} l Y Y Y Y @{}}
\toprule
\textbf{Loss} & \textbf{RMSE} & \textbf{MAE} & \textbf{R} & \textbf{CI} \\
\midrule
S1 & 1.249 & 0.938 & 0.820 & 0.816 \\
\textbf{S1+P} & \textbf{1.099} & \textbf{0.752} & \textbf{0.865} & \textbf{0.850} \\
\bottomrule
\end{tabularx}
\end{center}

Table \ref{tab:ablation_components} shows that hydrogen bond graph modeling substantially improves performance.

\begin{table*}[!t]
\centering
\tablesize
\caption{Ablation experiments on core components (PDBbind v2016 Core Set)}
\label{tab:ablation_components}
\begin{tabularx}{\linewidth}{@{} l Y Y Y Y @{}}
\toprule
\textbf{Configuration} & \textbf{RMSE} $\downarrow$ & \textbf{MAE} $\downarrow$ & \textbf{Pearson R} $\uparrow$ & \textbf{CI} $\uparrow$ \\
\midrule
SEQ & 1.959 & 1.576 & 0.497 & 0.672 \\
SEQ+SMILES & 1.429 & 1.146 & 0.759 & 0.778 \\
SEQ+Pocket & 1.539 & 1.220 & 0.707 & 0.753 \\
SEQ+Pocket+SMILES & 1.382 & 1.079 & 0.791 & 0.797 \\
SEQ+Pocket+SMILES+self-attention & 1.246 & 1.001 & 0.808 & 0.801 \\
SEQ+Pocket+SMILES+H-BondGNN & 1.118 & 0.851 & 0.848 & 0.838\\
\textbf{HBGSA} & \textbf{1.099} & \textbf{0.752} & \textbf{0.865} & \textbf{0.850} \\
\bottomrule
\end{tabularx}
\end{table*}

Table \ref{tab:ablation_lambda} presents the impact of different Pearson loss weight coefficients $\lambda$. Due to the long training time, we sampled seven representative values: $\lambda \in \{1, 25, 50, 75, 100, 125, 150\}$. The results reveal a U-shaped curve for RMSE and MAE: as $\lambda$ increases, prediction errors first decrease then increase, reaching optimal values at $\lambda=50$ (RMSE=1.099, MAE=0.752). Meanwhile, Pearson R and CI (measuring correlation and monotonicity respectively) improve monotonically with $\lambda$ up to $\lambda=75$, then plateau.

\begin{center}
\tablesize
\captionof{table}{Impact of Pearson loss weight $\lambda$}
\label{tab:ablation_lambda}
\begin{tabularx}{\columnwidth}{@{} l Y Y Y Y @{}}
\toprule
\textbf{$\lambda$} & \textbf{RMSE} & \textbf{MAE} & \textbf{R} & \textbf{CI} \\
\midrule
1 & 1.275 & 1.032 & 0.821 & 0.808 \\
25 & 1.212 & 0.881 & 0.840 & 0.829 \\
\textbf{50} & \textbf{1.099} & \textbf{0.752} & \textbf{0.865} & \textbf{0.850} \\
75 & 1.124 & 0.812 & 0.872 & 0.852 \\
100 & 1.163 & 0.855 & 0.869 & 0.849 \\
125 & 1.201 & 0.893 & 0.867 & 0.847 \\
150 & 1.253 & 0.944 & 0.864 & 0.844 \\
\bottomrule
\end{tabularx}
\end{center}

Figure \ref{fig:lambda_radar} visualizes these trade-offs using a radar chart, clearly showing that $\lambda=50$ achieves the most balanced performance across all metrics.

\begin{center}
\includegraphics[width=0.8\columnwidth]{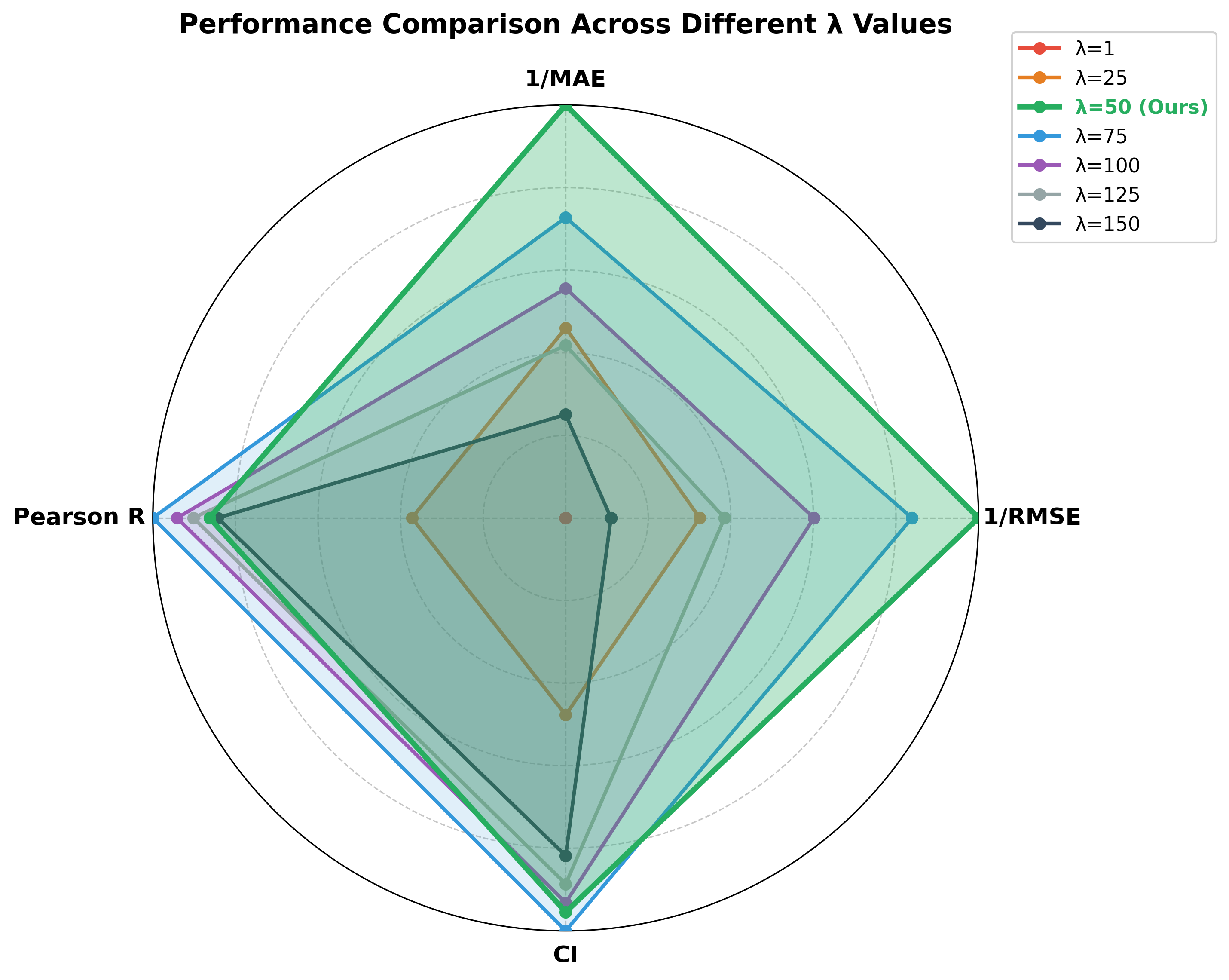}
\captionof{figure}{Radar chart of $\lambda$ trade-offs.}
\label{fig:lambda_radar}
\end{center}

To validate model stability, we conduct five-fold cross-validation. Table \ref{tab:ablation_cv} presents detailed results for each fold. The model maintains stable performance across different data partitions, with average RMSE of $1.110 \pm 0.028$, Pearson R of $0.864 \pm 0.007$, and CI of $0.851 \pm 0.006$. The small standard deviations validate the model's generalization capability and result reliability.

\subsection{Model Interpretability}
\justifying

\begin{figure*}[!t]
\centering
\begin{subfigure}[b]{0.48\textwidth}
\includegraphics[width=\textwidth]{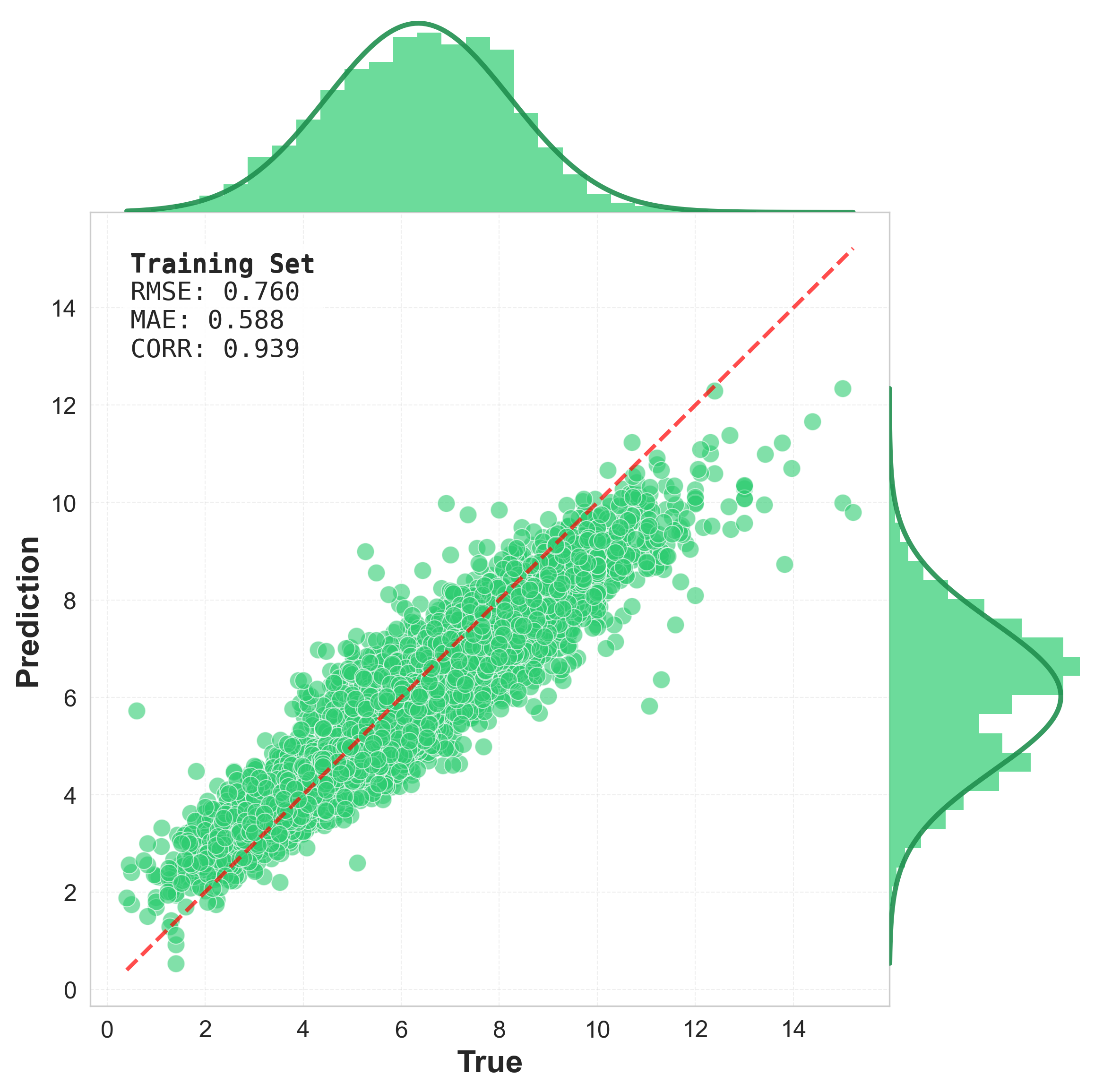}
\caption{Training Set}
\end{subfigure}
\hfill
\begin{subfigure}[b]{0.48\textwidth}
\includegraphics[width=\textwidth]{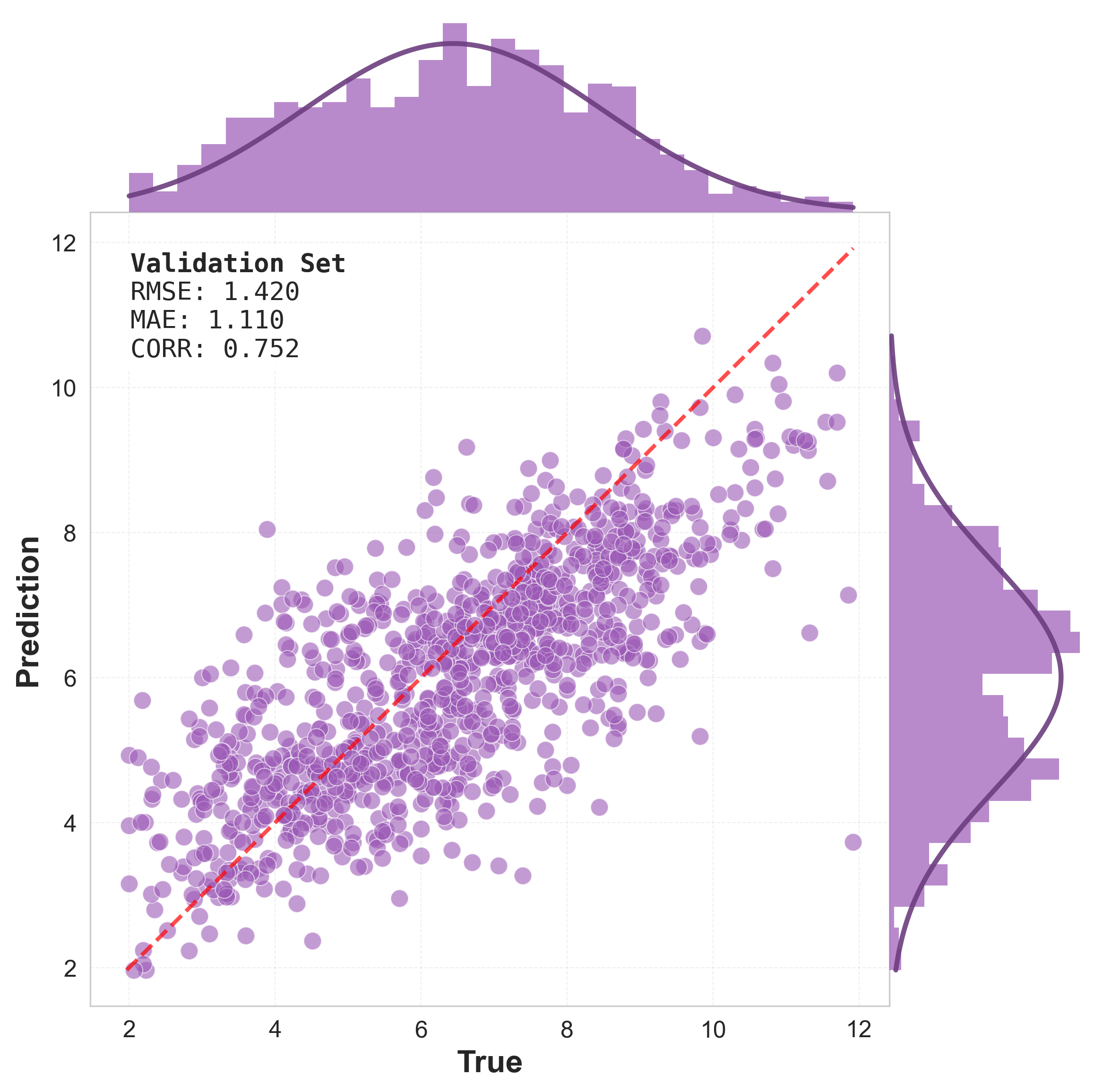}
\caption{Validation Set}
\end{subfigure}

\begin{subfigure}[b]{0.48\textwidth}
\includegraphics[width=\textwidth]{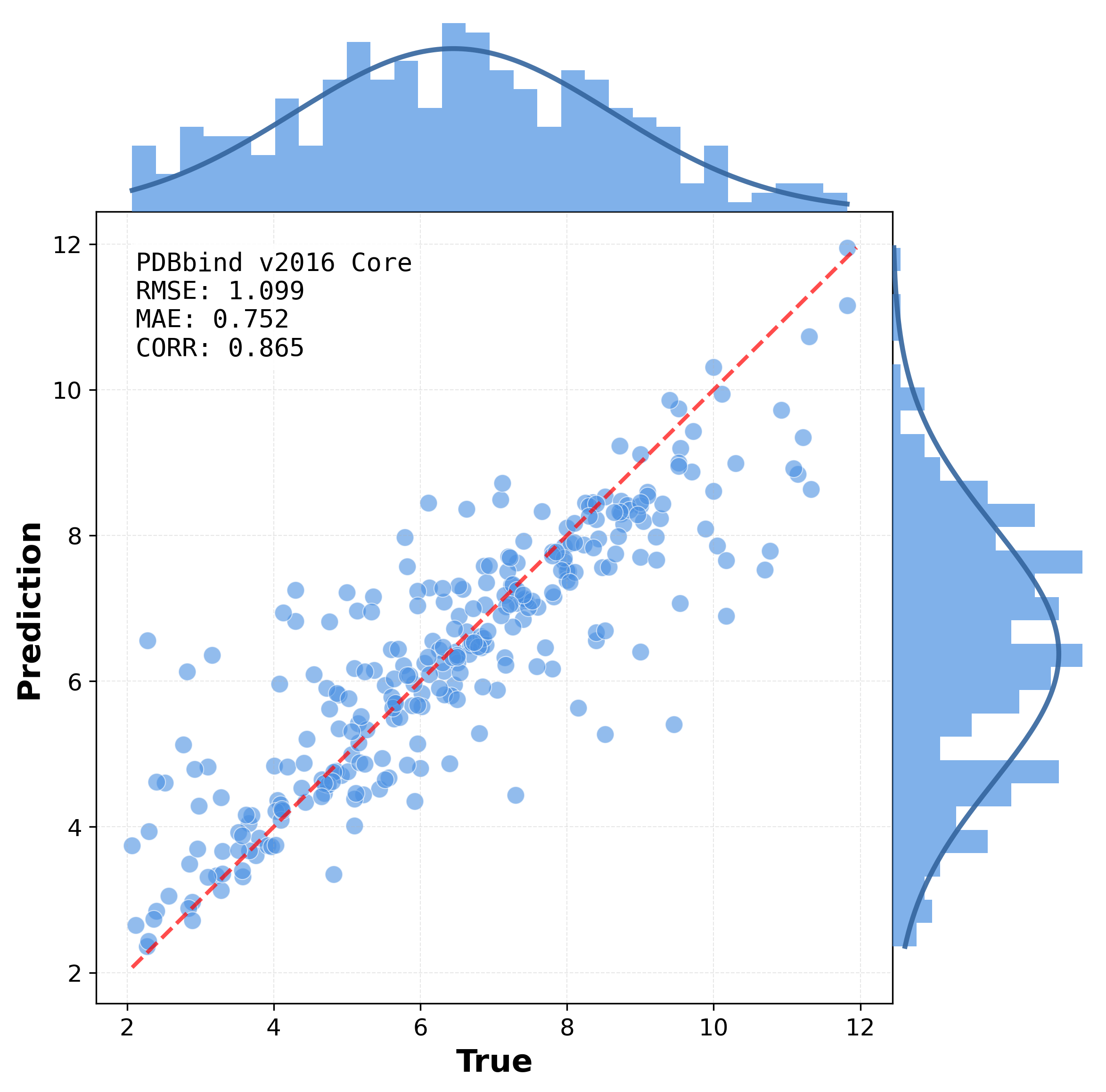}
\caption{Test Set}
\end{subfigure}
\hfill
\begin{subfigure}[b]{0.48\textwidth}
\includegraphics[width=\textwidth]{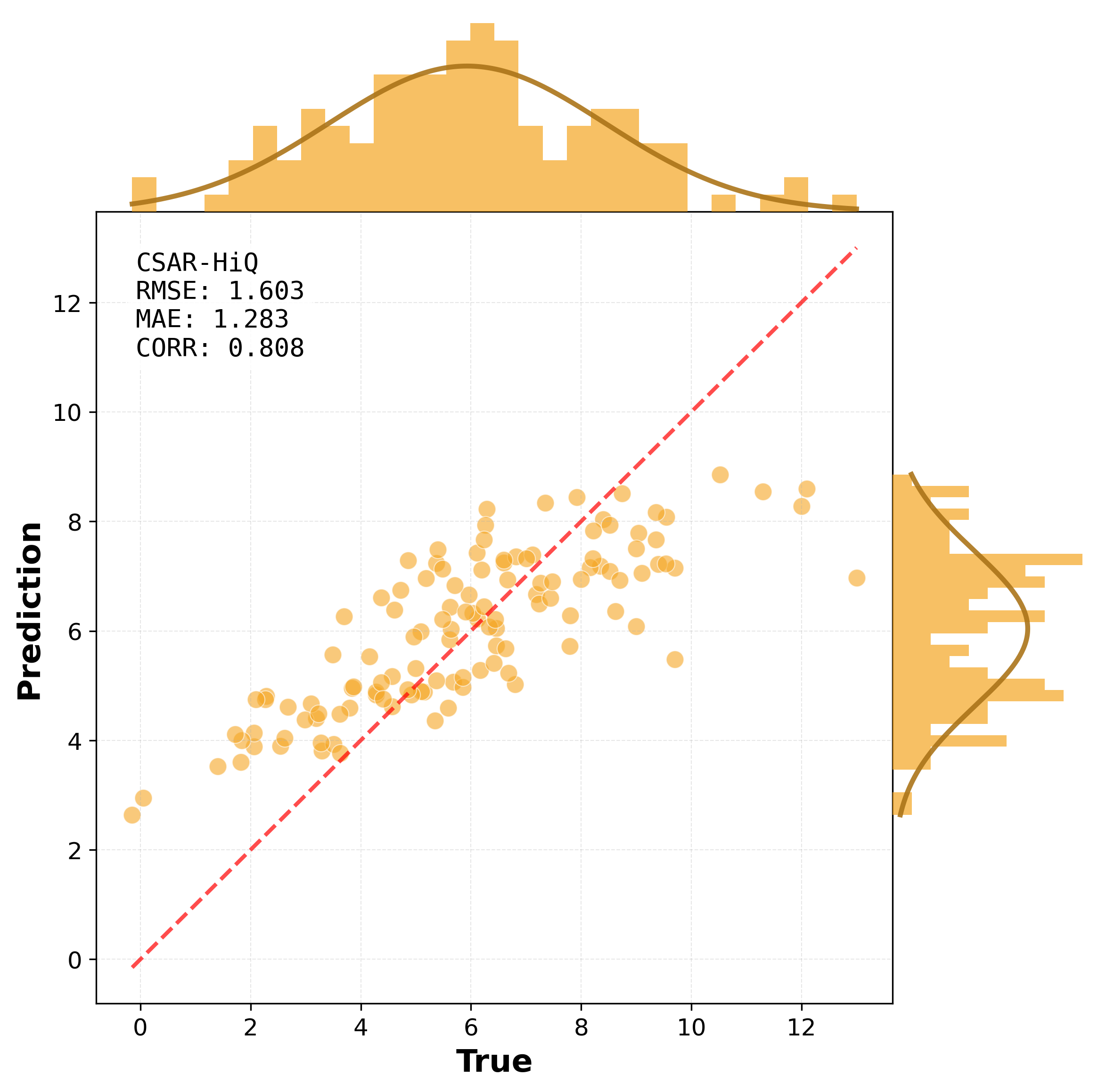}
\caption{CSAR-HiQ Dataset}
\end{subfigure}
\caption{Prediction performance on 4 sets.}
\label{fig:prediction_scatter_train_val}
\end{figure*}

To understand HBGSA's prediction mechanism, we conduct comprehensive visualization analysis across all data partitions. Figure \ref{fig:prediction_scatter_train_val} shows the model's prediction performance across all datasets. The model achieves strong performance on training and validation sets, with slight performance degradation on test set and CSAR-HiQ indicating good generalization without overfitting.

\begin{figure*}[!t]
\centering
\subcaptionbox{Training Set}{
\includegraphics[width=0.48\textwidth]{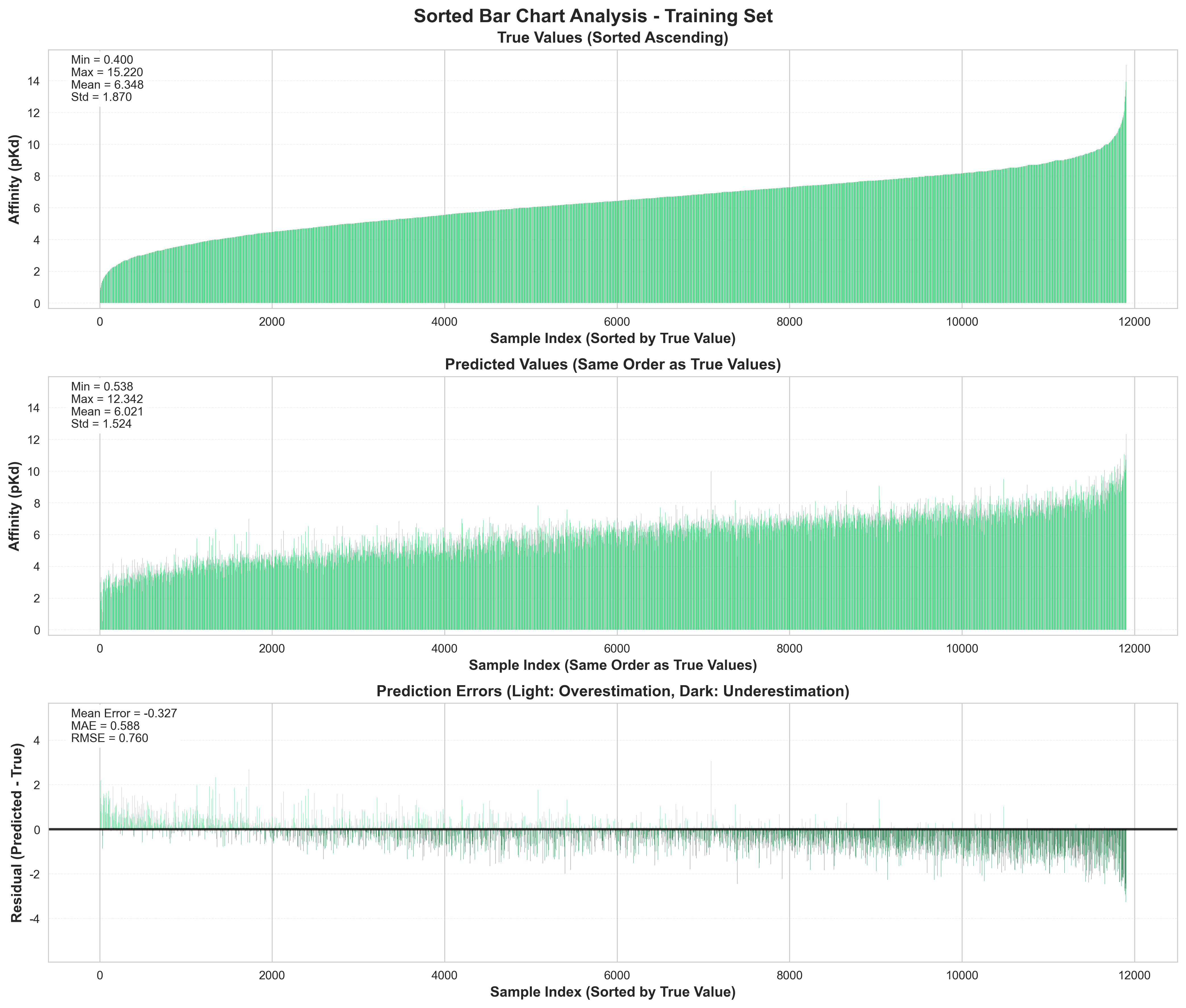}
}
\subcaptionbox{Validation Set}{
\includegraphics[width=0.48\textwidth]{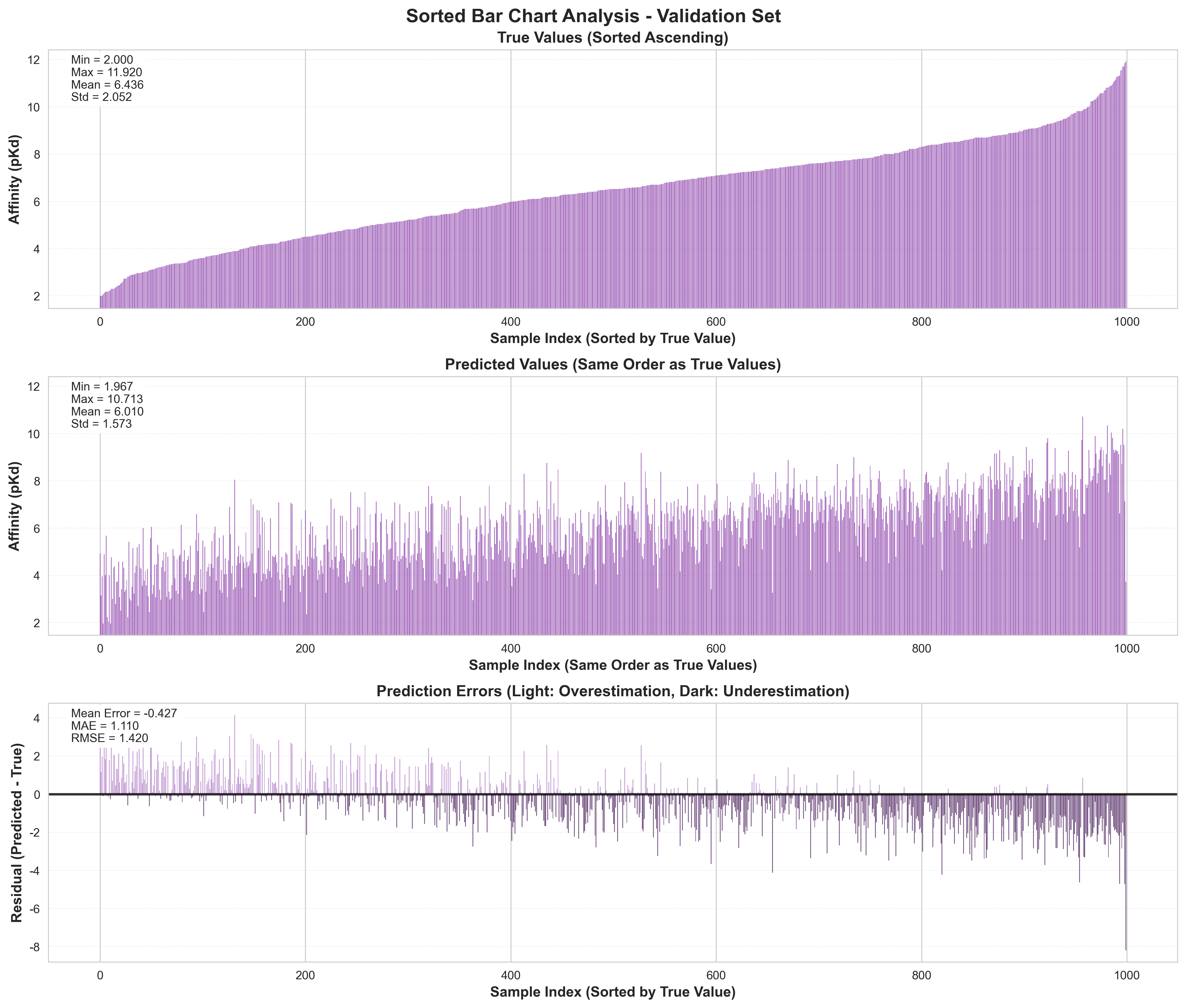}
}
\subcaptionbox{Test Set}{
\includegraphics[width=0.48\textwidth]{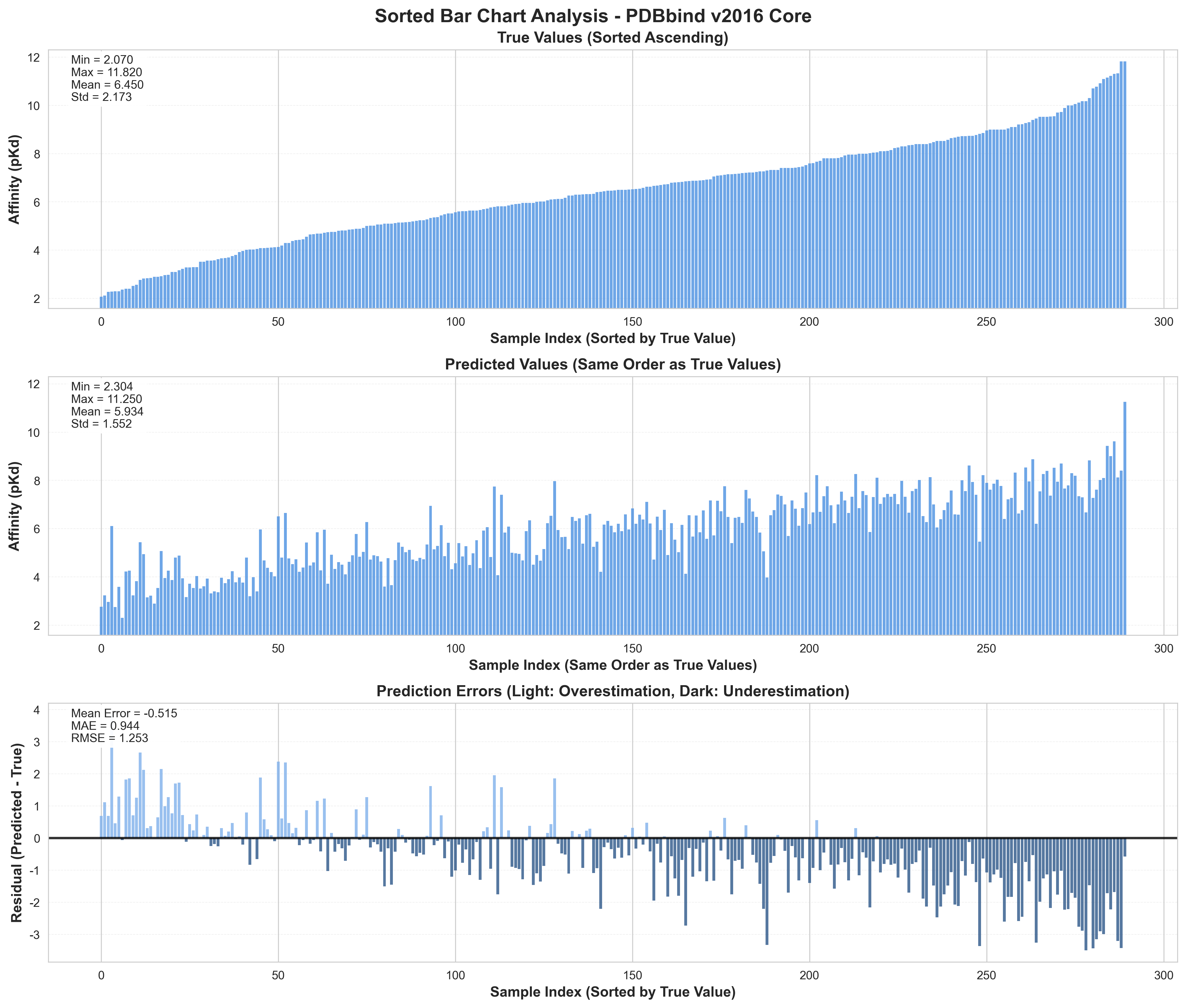}
}
\subcaptionbox{CSAR-HiQ Dataset}{
\includegraphics[width=0.48\textwidth]{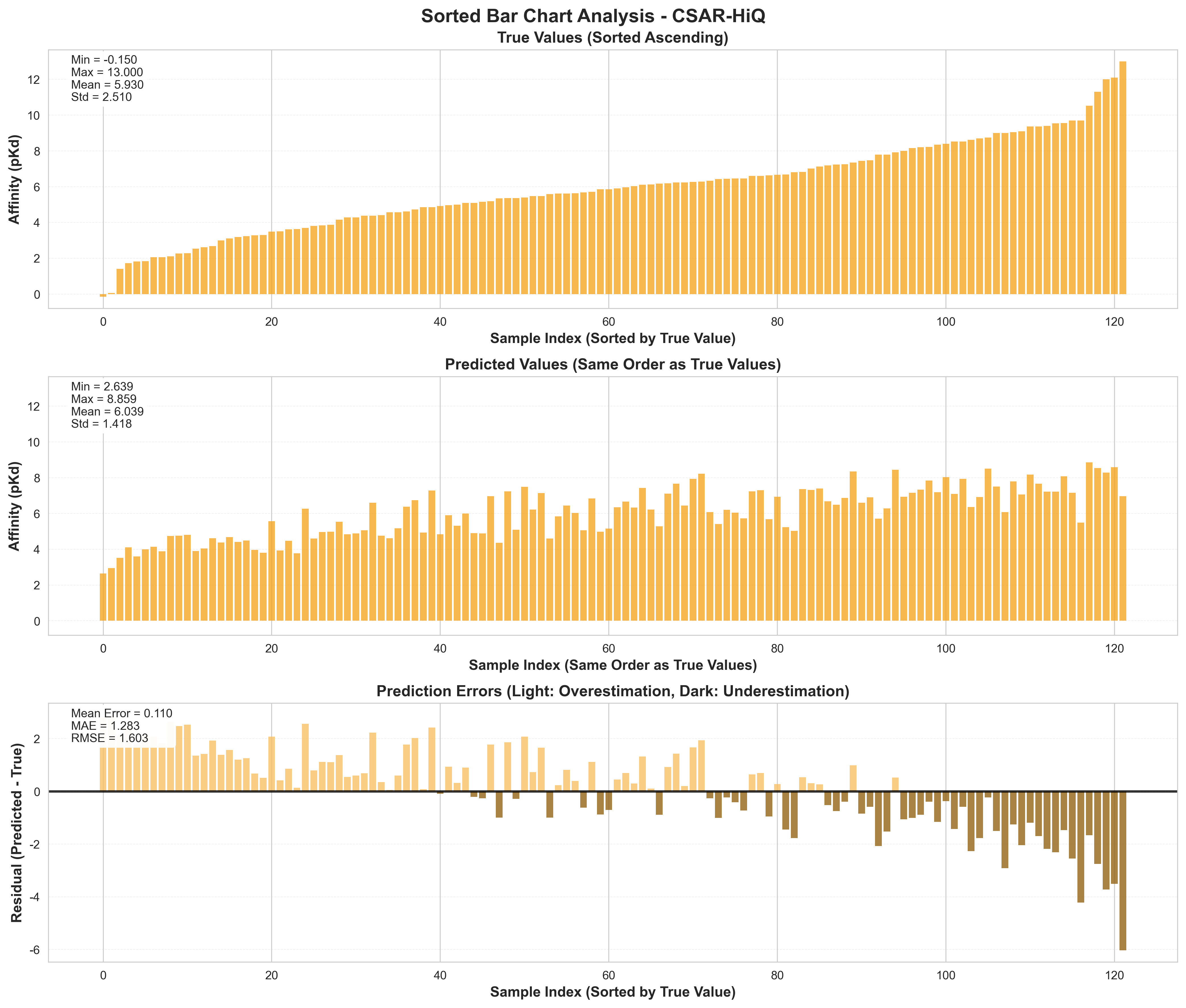}
}
\caption{Sorted bar chart analysis for 4 sets.}
\label{fig:sorted_bar_train_val}
\end{figure*}

To visualize the model's prediction monotonicity and correlation quality, we present sorted bar chart analysis across all datasets. Figure \ref{fig:sorted_bar_train_val} shows sorted bar charts for all four datasets (training, validation, test, and CSAR-HiQ), where samples are sorted by true affinity values. Strong correlation manifests as smooth monotonic trends in predicted values.

HBGSA maintains strong monotonicity across all datasets, as evidenced by the smooth trends in sorted bar charts. This monotonicity reflects the high correlation between predictions and true affinities, which is the direct result of Pearson loss optimization. The improved CI values (measuring monotonicity) stem from enhanced correlation.

\subsection{Hydrogen Bond Density Analysis}
\justifying

Statistical analysis of the training dataset reveals that the hydrogen bond count distribution supports our choice of $N=20$ (Figure \ref{fig:hbond_count_distribution}). The average number of hydrogen bonds is 8.22 (median=7, std=5.67). The distribution shows that 96.61\% of samples contain 20 or fewer hydrogen bonds, with only 3.39\% exceeding this threshold. The 95th percentile is 19 bonds, confirming that $N=20$ provides comprehensive coverage while maintaining computational efficiency.
\vspace{1em}
\vspace{1em}
\vspace{1em}
\vspace{1em}

\begin{table*}[!t]
\centering
\renewcommand{\arraystretch}{0.75}
\tablesize
\caption{Five-fold cross-validation results}
\label{tab:ablation_cv}
\begin{tabularx}{\linewidth}{@{} l Y Y Y Y @{}}
\toprule
Fold & RMSE $\downarrow$ & MAE $\downarrow$ & Pearson R $\uparrow$ & CI $\uparrow$ \\
\midrule
Fold 1  & 1.1410 & 0.8124 & 0.8587 & 0.8436 \\
Fold 2  & 1.0691 & 0.7529 & 0.8769 & 0.8595 \\
Fold 3  & 1.1200 & 0.7394 & 0.8581 & 0.8529 \\
Fold 4  & 1.1143 & 0.7675 & 0.8607 & 0.8464 \\
Fold 5  & 1.1054 & 0.7396 & 0.8656 & 0.8514 \\
\midrule
Mean $\pm$ Std & $1.110 \pm 0.028$ & $0.762 \pm 0.036$ & $0.864 \pm 0.007$ & $0.851 \pm 0.006$ \\
\bottomrule
\end{tabularx}
\end{table*}

\begin{center}
\includegraphics[width=\columnwidth]{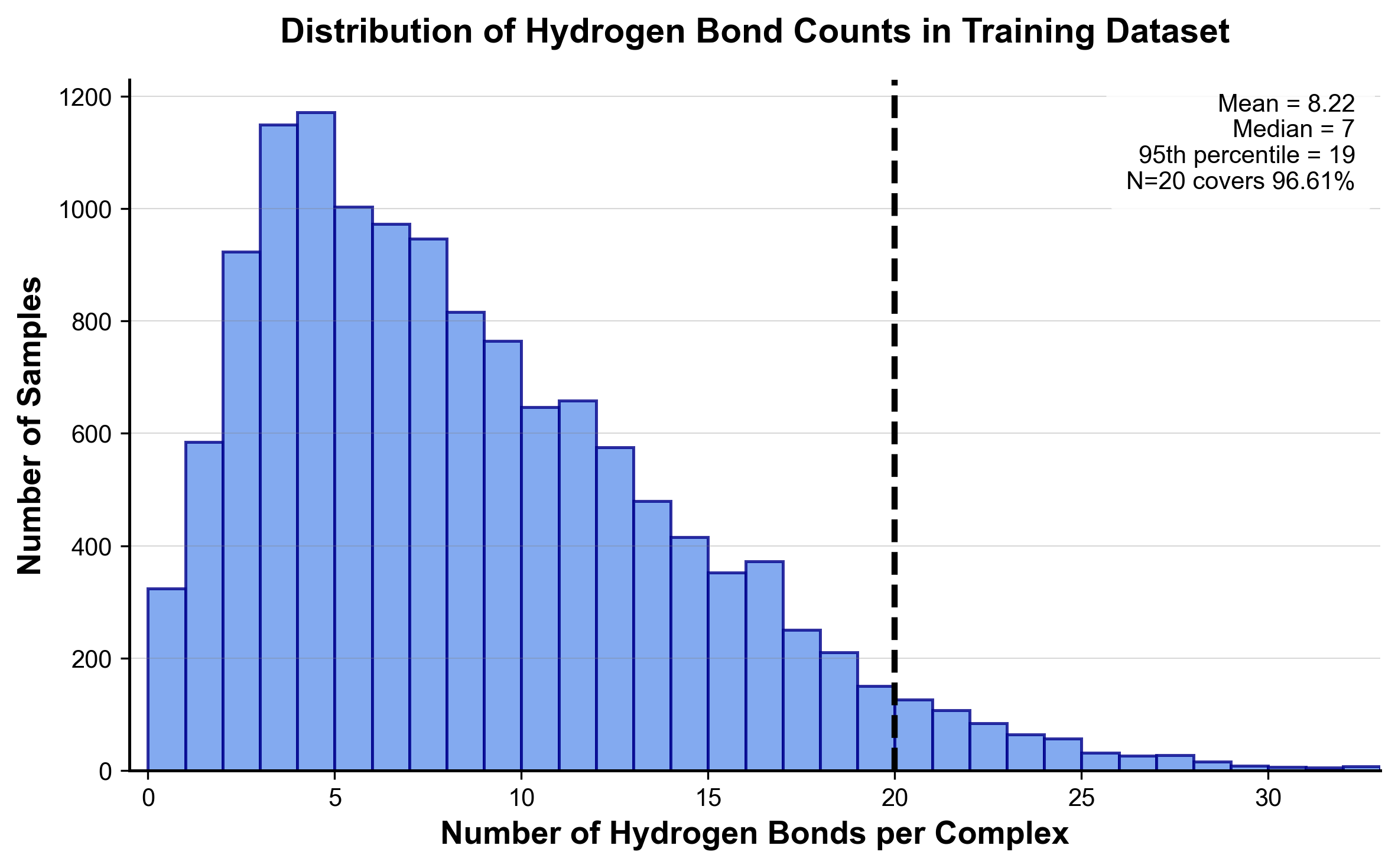}
\captionof{figure}{Distribution of hydrogen bond counts in the training dataset. }
\label{fig:hbond_count_distribution}
\end{center}

\vspace{1em}
\vspace{1em}

\begin{figure*}[t!]
\centering
\subcaptionbox{High-affinity complex (pdbid: 1zsb, $K_d=0.6$ nM)}{
\includegraphics[width=0.48\textwidth]{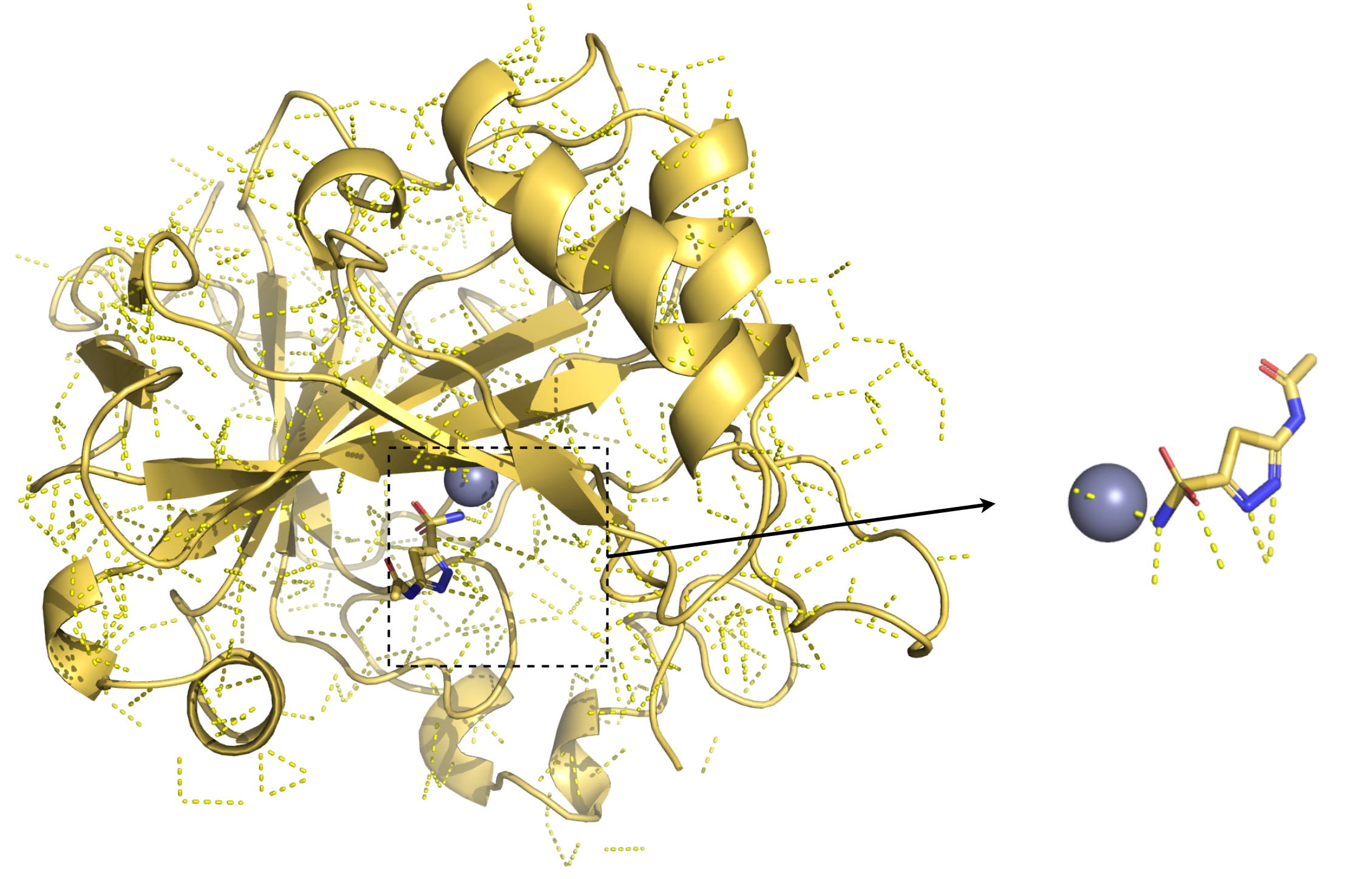}
}
\subcaptionbox{Low-affinity complex (pdbid: 3ekv, $K_d=9.41$ nM)}{
\includegraphics[width=0.48\textwidth]{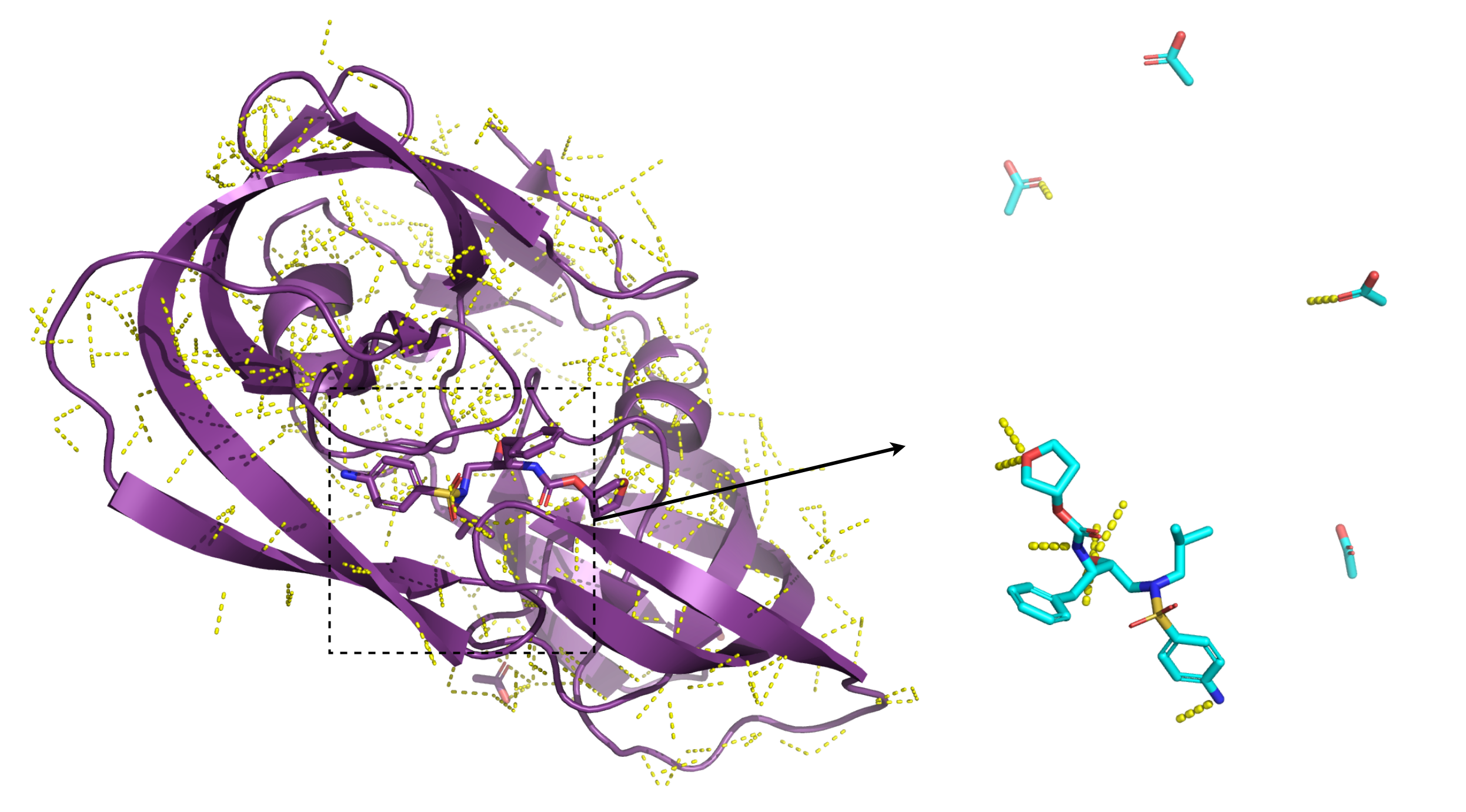}
}
\caption{Relationship between hydrogen bond density and binding affinity. Yellow dashed lines indicate hydrogen bonds.}
\label{fig:hbond_density_analysis}
\end{figure*}

To validate the physical basis of HBGSA's performance improvements, we analyze the relationship between hydrogen bond density and binding affinity using representative complexes from the PDBbind dataset. Figure \ref{fig:hbond_density_analysis} presents two typical cases, revealing the quantitative relationship between hydrogen bond density and binding affinity.

Hydrogen bond density, rather than absolute count, fundamentally determines binding affinity. Both complexes use acetazolamide (AZM) but differ significantly in affinity: 1zsb (5 bonds, 19 atoms, 26.3\% density, $K_d=0.6$ nM) vs. 3ekv (9 bonds, 57 atoms, 15.8\% density, $K_d=9.41$ nM). The H94N mutation in 3ekv disrupts $Zn^{2+}$ chelation, destabilizing the metal center despite more hydrogen bonds. This confirms that affinity depends on density, geometry, and structural environment—not bond count alone.

We define hydrogen bond density as the ratio of hydrogen bonds to ligand atoms:
\begin{equation}
\rho_{hbond} = \frac{N_{hbond}}{N_{ligand}}
\end{equation}
where $N_{hbond}$ is the number of hydrogen bonds formed and $N_{ligand}$ is the number of ligand atoms. Table \ref{tab:hbond_density_details} presents detailed quantitative analysis for both complexes, showing that despite having fewer hydrogen bonds (5 vs. 9), complex 1zsb achieves higher affinity due to its superior hydrogen bond density (26.3\% vs. 15.8\%).

\vspace{1em}
\vspace{1em}
\begin{center}
\small
\captionof{table}{Hydrogen bond density analysis in representative complexes}
\label{tab:hbond_density_details}
\begin{tabularx}{\columnwidth}{@{} l Y Y @{}}
\toprule
\textbf{Property} & \textbf{1zsb} & \textbf{3ekv} \\
\midrule
Ligand &Acetazolamide&Acetazolamide\\
Formula & {$C_4H_6N_4O_3S_2$}&{$C_4H_6N_4O_3S_2$} \\
Molecules & 1 & 3 \\
$N_{ligand}$ & 19 & 57 \\
$N_{hbond}$ & 5 & 9 \\
$\rho_{hbond}$ & 26.3\% & 15.8\% \\
$K_d$ (nM) & 0.6 & 9.41 \\
Affinity & high & low \\
\bottomrule
\end{tabularx}
\end{center}

\vspace{0.5em}

\section{Discussion and Conclusion}
\justifying

This section discusses the underlying physical mechanisms behind these improvements, implications for virtual screening workflows, and current model limitations.

\subsection{Physical Mechanism of Hydrogen Bond Graph Architecture}
\justifying

HBGSA models hydrogen bonds through graph neural networks, capturing spatial topology and cooperative effects more effectively than distance-threshold methods.

\subsection{Value of Correlation-Based Loss in Virtual Screening}
\justifying

Virtual screening aims to identify promising drug candidates from large compound libraries (typically millions of compounds) for experimental validation. In this context, prediction-target correlation is critical because virtual screening decisions are based on relative comparisons rather than absolute values. For instance, researchers do not need to know whether a compound's binding affinity is precisely 5.23 nM or 5.45 nM; instead, they need to determine which compounds are most likely to exhibit high affinity.

Correlation-based optimization offers three key advantages. First, it is naturally robust to systematic bias. Pearson correlation is invariant to linear transformations of predictions—if all predictions are systematically shifted by a constant or scaled by a factor, the correlation remains unchanged. Second, correlation-based optimization encourages models to maintain correct relative relationships, even if absolute errors are larger. This property is crucial for identifying the small subset of high-affinity compounds that are the primary targets of virtual screening. Third, by directly optimizing the linear relationship between predictions and true values, correlation-based loss preserves the model's ability to distinguish high-affinity from low-affinity compounds across the full range of binding affinities.

Our hybrid loss function (SmoothL1 + Pearson) balances these considerations: SmoothL1 loss ensures reasonable absolute accuracy, while Pearson loss optimizes correlation to maintain correct relative relationships. The ablation study (Table \ref{tab:ablation_lambda}) demonstrates this balance—as $\lambda$ increases from 1 to 50, correlation improves (Pearson R: 0.821→0.865), and as a natural consequence, prediction monotonicity improves (CI: 0.808→0.850). However, excessive $\lambda$ values create a trade-off—while $\lambda=75$ achieves the highest correlation (Pearson R=0.872), absolute prediction accuracy degrades with RMSE increasing by 2.3\% and MAE by 8.0\% compared to $\lambda=50$. The optimal $\lambda=50$ achieves both strong correlation and acceptable absolute accuracy, making the model particularly suitable for virtual screening applications.

\subsection{Model Limitations}
\justifying

HBGSA relies on high-quality static crystal structures, which may not capture conformational flexibility in physiological conditions. For example, kinase DFG-flip (Asp-Phe-Gly motif transition) dramatically alters affinity but cannot be captured by static structure models. Future work will explore conformational ensemble modeling to address this limitation.

\subsection{Conclusion}

HBGSA demonstrates that explicit modeling of hydrogen bond spatial features through graph neural networks, combined with correlation-optimized loss, significantly improves both prediction accuracy and cross-dataset generalization in drug-target affinity prediction. The correlation-based optimization naturally enhances prediction monotonicity, making the model particularly suitable for virtual screening applications. The model achieves state-of-the-art performance on PDBbind Core Set while maintaining robust generalization on CSAR-HiQ.

Future work will explore conformational flexibility modeling and model compression techniques for efficient deployment in large-scale virtual screening.

\vspace{1em}

\noindent
\fbox{\begin{minipage}{0.95\columnwidth}
\textbf{Key Points}

\begin{itemize}
\item We propose HBGSA, which integrates three key innovations: (1) hydrogen bond graph neural networks that explicitly model spatial topology and cooperative effects of hydrogen bonds, (2) self-attention mechanisms on sequence encoders to enhance feature representation, and (3) Pearson correlation loss to optimize prediction-target correlation.

\item HBGSA achieves state-of-the-art performance on PDBbind Core Set (RMSE=1.099, Pearson R=0.865) and demonstrates strong generalization on CSAR-HiQ dataset (RMSE=1.603, Pearson R=0.808).

\item Ablation studies confirm that all three components—hydrogen bond GNN, self-attention, and Pearson correlation loss—contribute significantly to performance improvements.
\end{itemize}
\end{minipage}}

\vspace{1em}

\section*{Author Contributions}

 Junxiao Kong (Co-first author): Conceptualization, Methodology, Software, Investigation, Writing - Original Draft. Chupei Tang (Co-first author): Data Curation, Validation (Baseline Experiments), Methodology Optimization. Di Wang: Methodology Optimization, Software Testing. Jixiu Zhai: Validation, Writing - Review \& Editing. Yi He: Resources, Validation. Moyu Tang: Resources, Validation. Tianchi Lu (Corresponding author):Methodology, Supervision, Project Administration, Funding Acquisition, Writing - Review \& Editing (Final Version).

\section*{Acknowledgments}

We thank all members of our research group for their valuable discussions and feedback throughout this project.

\section*{Competing interests}

No competing interest is declared.

\section*{Data Availability}

The original data source is from \url{https://www.pdbbind-plus.org.cn/}. Supplementary materials including detailed hyperparameter configurations, parameter breakdown, and evaluation metric calculations are provided in the Supplementary File.


\end{multicols}

\end{document}